\documentclass[10pt,twocolumn,letterpaper]{article}

\usepackage[pagenumbers]{cvpr} 

\usepackage{graphicx}
\usepackage{amsmath}
\usepackage{amssymb}
\usepackage{booktabs}

\usepackage{multicol}
\usepackage{multirow}
\usepackage[table]{xcolor}
\usepackage{makecell}
\usepackage{pifont}
\newcommand{\cmark}{\ding{51}}

\usepackage[pagebackref,breaklinks,colorlinks]{hyperref}

\usepackage[capitalize]{cleveref}
\crefname{section}{Sec.}{Secs.}
\Crefname{section}{Section}{Sections}
\Crefname{table}{Table}{Tables}
\crefname{table}{Tab.}{Tabs.}

\def\cvprPaperID{*****} 
\def\confName{CVPR}
\def\confYear{2022}

\begin{document}

\title{Monocular 3D Hand Pose Estimation with Implicit Camera Alignment}


\author{
Christos Pantazopoulos\textsuperscript{1}\thanks{This work is part of the author's diploma thesis} \quad
Spyridon Thermos\textsuperscript{2} \quad
Gerasimos Potamianos\textsuperscript{1} \\[1ex]
\textsuperscript{1}Department of Electrical and Computer Engineering, University of Thessaly, Greece \\
\textsuperscript{2}Moverse \\
\texttt{cpantazop@uth.gr} \quad \texttt{spiros@moverse.ai}\quad \texttt{gpotam@ieee.org}
}
\maketitle

\begin{abstract}
Estimating the 3D hand articulation from a single color image is an important problem with applications in Augmented Reality (AR), Virtual Reality (VR), Human-Computer Interaction (HCI), and robotics. 
Apart from the absence of depth information, occlusions, articulation complexity, and the need for camera parameters knowledge pose additional challenges. In this work, we propose an optimization pipeline for estimating the 3D hand articulation from 2D keypoint input, which includes a keypoint alignment step and a fingertip loss to overcome the need to know or estimate the camera parameters.
We evaluate our approach on the EgoDexter and Dexter+Object benchmarks to showcase that it performs competitively with the state-of-the-art, while also demonstrating its robustness when processing ``in-the-wild" images without any prior camera knowledge. Our quantitative analysis highlights the sensitivity of the 2D keypoint estimation accuracy, despite the use of hand priors. Code is available at the project page \url{https://cpantazop.github.io/HandRepo/}

\end{abstract}

\section{Introduction}
\label{sec:intro}

Reconstructing an articulated 3D hand from a single RGB image is a challenging problem in computer vision with a wide range of applications, including augmented and virtual reality (AR/VR), human-computer interaction (HCI), robotics, and the metaverse. 
However, it is a challenging task due to the lack of depth information, frequent object-related occlusions and self-occlusions, unknown camera intrinsics parameters, and the hand’s complex articulation.

In this paper, we propose an alternative method for monocular 3D hand pose estimation that operates without prior camera parameters knowledge. Our approach leverages the robust 2D keypoint detection capabilities of Media\-Pipe~\cite{Mediapipehand}, combined with a two-stage optimization pipeline that fits the MANO~\cite{MANO} hand model to the detected 2D keypoints. The first stage performs a rigid transformation to align the initial MANO hand model with the 2D detections, establishing a coarse 3D pose estimation. The second stage refines this estimate using a fingertip alignment loss and anatomical constraints to ensure physically plausible hand configurations.

By avoiding reliance on known camera parameters, our method is able to perform in-the-wild while maintaining accuracy. The anatomical constraints operate as a regularizer, preventing unrealistic hand poses, while the fingertip alignment loss improves precision in critical regions. We demonstrate that our approach achieves competitive performance compared to state-of-the-art (SotA) methods, even without knowing the camera intrinsic parameters, making it a practical solution for real-world applications.

Our key contributions include:
\begin{enumerate}
    \item A camera-agnostic 3D hand pose estimation framework that leverages 2D keypoint detections.
    \item A two-stage optimization pipeline combining rigid alignment and refinement with anatomical and fingertip constraints.
    \item An extensive evaluation showcasing the method's robustness in-the-wild and its competitive performance on standard benchmarks.
\end{enumerate}

The rest of the paper is organized as follows: a) the prior work in 2D and 3D hand pose estimation and the relevant parametric models are discussed in Sec.~\ref{sec:prior}; a more detailed background of the leveraged modules is presented in Sec.~\ref{sec:back}; our method is detailed in Sec.~\ref{sec:method}; the experimental framework is presented in Sec.~\ref{sec:experiment}; and finally our conclusions are reported in Sec.~\ref{sec:conclusion}

\begin{figure}[!t]
  \centering
   \includegraphics[width=0.28\linewidth]{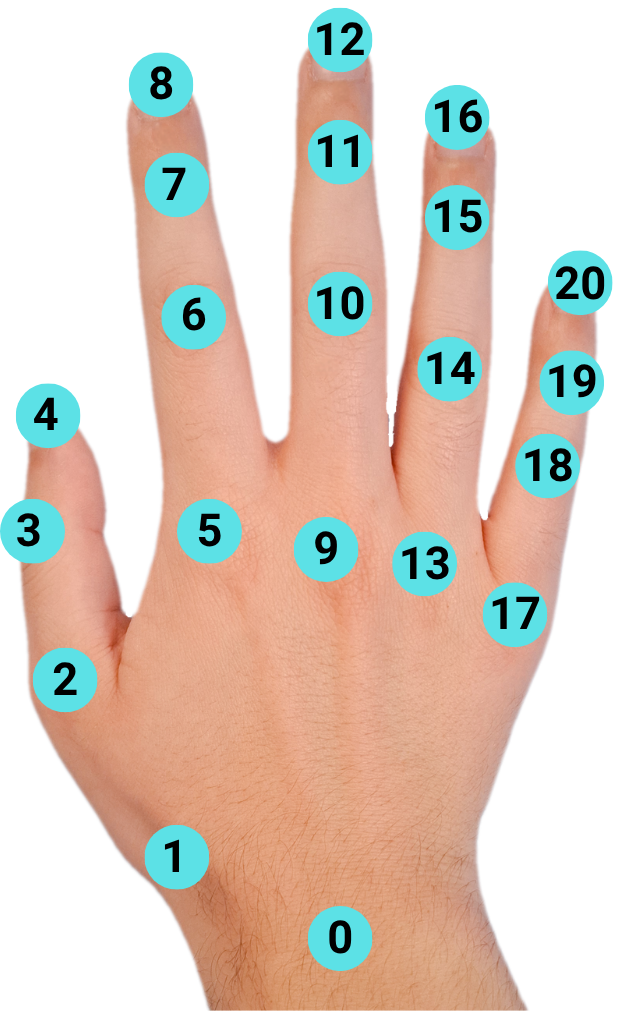}

   \caption{The 21 hand keypoints estimated by MediaPipe~\cite{Mediapipehand}.}
   \label{bloghands}
\end{figure}

\section{Prior Art}
\label{sec:prior}
Hand pose estimation refers to predicting the position and orientation of the hand and fingers in relation to a set coordinate system using either RGB images, volumetric data from depth cameras, or a combination of both.
This paper focuses on implementations that exploit solely the color information of the human hand.

\textbf{2D hand pose estimation.}
In 2D hand pose estimation, to estimate the pose of the hand we need to estimate the location of its keypoints. 
The human hand has 21 keypoints \cite{Blog}: 
In Fig.~\ref{bloghands}  we depict the keypoints on a hand. 
In prior work on this topic, in \cite{simon2017hand} Simon~\etal present an approach that uses a multi-camera system to train fine-grained detectors for keypoints that are prone to occlusion, such as the joints of a hand.
This procedure is called multiview bootstrapping.
It uses an initial keypoint detector to generate noisy labels across multiple views, triangulates valid detections in 3D, and reprojects them as new training data to iteratively improve the detector. This process yields a real-time RGB keypoint detector with accuracy comparable to depth-based methods.

\textbf{3D Hand Pose Estimation.}
There have been numerous methods estimating 3D pose using depth or multi-view sensors.
However, regressing pose from a single RGB image is challenging due to the fact that 3D pose requires some form of depth estimates, which are ambiguous given only an RGB image. 
This was later dealt with the use of parametric hand models like we do in our proposed method.
Iqbal~\etal \cite{iqbal2018hand} propose a new method for 3D hand pose estimation from a monocular image through a 2.5D pose representation. 
Zimmermann and Brox \cite{zimmermann2017learning} also present an approach that estimates 3D hand pose from RGB images. To handle the ``missing" depth data they propose a deep network that learns a network-implicit 3D articulation prior. Together with detected keypoints in the images, this network yields good estimates of the 3D pose.
Additionally, Zimmermann~\etal \cite{zimmermann2019freihand} introduce a large-scale 3D hand pose dataset based on synthetic hand models for training the involved networks.

\textbf{3D Hand Pose Estimation using Parametric Hand Models.}
A stepping stone to the evolution of the 3D hand pose estimation from a single RGB image without the use of depth information or other cameras/sensors has been the introduction of hand parametric models like MANO \cite{MANO}.

Boukhayma~\etal \cite{boukhayma20193d} present the first end-to-end deep learning based method that predicts both 3D hand shape and pose from RGB images in the wild. This network consists of the concatenation of a deep convolutional encoder and a fixed model-based decoder. 
Panteleris~\etal \cite{panteleris2017using} present a method for the real-time estimation of the full 3D pose of one or more human hands using a single commodity RGB camera. More specifically, given an RGB image and the relevant camera calibration information, they employ a SotA detector to localize hands. Then, using a crop of a hand in the image, they run the pretrained network of OpenPose \cite{simon2017hand} for hands to estimate the 2D location of hand joints. Finally, non-linear least-squares minimization fits a 3D model of the hand, distinct from the MANO model, to the estimated 2D joint positions, recovering the 3D hand pose.
Mueller~\etal \cite{GANeratedHands_CVPR2018} address the problem of real-time 3D hand tracking based on a monocular RGB-only sequence. 
Their method combines a CNN with a kinematic 3D hand model. 
For training this CNN they generated a synthetic training dataset by using a neural network that translates synthetic
images to ``real" images, such that the so-generated images follow the same statistical distribution as real-world
hand images.
Ge~\etal \cite{ge20193d} propose a Graph CNN based method to reconstruct a full 3D mesh of hand surface that contains richer information of both 3D hand shape and pose. 
Baek~\etal \cite{baek2019pushing} also adopt the MANO parametric 3D hand model. To achieve the model fitting to RGB images they implement a hand mesh estimator by a neural network and a differentiable renderer, supervised by 2D segmentation masks and 3D skeletons.

Kulon~\etal \cite{kulon2020weaklysupervised} introduce a simple and effective network architecture for monocular 3D hand pose estimation consisting of an image encoder followed by a mesh convolutional decoder that is trained through a direct 3D hand mesh reconstruction loss. They train the network by gathering a large-scale dataset of hand action in YouTube videos and use it as a source of weak supervision.
Zhang~\etal \cite{zhang2019endtoend} present a Hand Mesh Recovery framework to tackle the problem of reconstructing the full 3D mesh of a human hand from a single RGB image.
The mesh representation is achieved by parameterizing MANO. To this end, a differentiable re-projection loss is defined in terms of the derived MANO representations and the ground-truth labels, thus making this framework end-to-end trainable.
Drosakis and Argyros \cite{inproceedings} present a method for simultaneous 3D hand shape and pose estimation on a single RGB image frame. Specifically, their method fits
the MANO 3D hand model to 2D hand keypoints, based on a 2D objective function that exploits anatomical joint limits, combined with shape regularization.

Lim~\etal \cite{mobilehand} present an approach for real-time estimation of 3D hand shape and pose from a single RGB image, using an efficient CNN named MobileNetV3-Small to extract key features from an input image. The extracted features are then sent to an iterative 3D regression module to infer camera parameters, hand shapes, and joint angles for projecting and articulating a 3D hand model.

\section{Background}
\label{sec:back}
\textbf{MANO - A 3D Hand Parametric Model.}

MANO is a parametric model commonly used in computer vision and graphics to encode the shape and pose variations of human hands. 
The MANO hand model takes as input 45 rotation parameters \(\theta\) and 10 shape parameters \(\beta\) to produce a 3D hand mesh. Once the model - denoted as $\Phi$ - has been loaded properly, it can be initialized by defining the following variables: 
\begin{equation}
    \Phi\{\beta, \theta, r\} \quad \mathrm{with} \quad \beta \in \mathbb{R}^{10}, \theta \in \mathbb{R}^{45}, r \in \mathbb{R}^{3},
\end{equation}
\noindent which correspond to the beta, pose, and global orientation parameters, respectively.
The shape parameters, denoted as \( \beta \), control the overall structure of the hand, such as the width of the fingers and palm. These parameters are derived through PCA on real hand scans, with only 10 parameters required to represent nearly all human hand shape variations. 
On the other hand, the pose parameters, denoted as \( \theta \), define the rotations of the hand's joints, allowing for various articulated poses. These rotations are represented in an axis-angle format for the 15 major joints of the hand, leading to a total of 45 values.
Each model implementing MANO hand has different but similar functions available to obtain the joints, faces and vertices of the mesh in order to later visualize it.

Our implementation is based on MANOTorch \cite{manotorch}, a differentiable PyTorch layer that deterministically maps MANO's pose and shape parameters to hand joints and vertices using PyTorch. 

\textbf{MediaPipe.}  
The first step in our method is to detect and localize the 21 keypoints of the hand in the input image. This can be achieved using specialized frameworks that serve as keypoint encoders, such as Media\-Pipe \cite{Mediapipehand}, OpenPose \cite{OpenPose}, and MMPose \cite{MMPose}. We chose to use the Media\-Pipe Hand Landmarker \cite{lugaresi2019mediapipe} that takes image data as input and outputs hand landmarks in image coordinates, hand landmarks in world coordinates, and handedness (left/right hand) of multiple detected hands.
The keypoint coordinates provided by Media\-Pipe are normalized and scaled between 0 and 1, so we denormalize them using the width and height of the image.

\textbf{Optimization Methods.}
The BFGS (Broyden-Fletcher-Goldfarb-Shanno) algorithm is an iterative optimization method used to solve unconstrained optimization problems \cite{num_optimization_2006}. 
Limited-memory BFGS (L-BFGS) \cite{Nocedal1980UpdatingQM} is a Quasi-Newton optimization method that builds upon the BFGS algorithm while significantly reducing memory usage.

\textbf{Loss Functions.}
The Mean Squared Error (MSE) \cite{probcourse} loss computes the average squared difference between the predicted values $\hat{y}_i$ and the true values $y_i$:
\begin{equation}
\text{MSE} = \frac{1}{n} \sum_{i=1}^{n} (y_i - \hat{y}_i)^2
\end{equation}
where $n$ is the number of data points.

The Geman-McClure (GM) loss function is a robust alternative to the MSE loss, designed to mitigate the influence of outliers. It is defined as:
\begin{equation}
L_{\text{GM}}(r) = \frac{\rho^2 r^2}{r^2 + \rho^2},
\end{equation}
where \( r \) represents the residual error (\textit{i.e.}, the difference between predicted and true values), and \( \rho \) is a parameter that controls the sensitivity of the function to large residuals.
It was originally introduced in the context of tomographic image reconstruction \cite{Geman1987StatisticalMF}, and has since been applied in various domains, including the 3D human pose estimation \cite{pavlakos2019expressivebodycapture3d}.

The Huber loss \cite{Huber1964RobustEO} combines the MSE and MAE behavior and is defined as:

\begin{equation}
L_{\delta}(a) =
\begin{cases} 
\frac{1}{2}\:\ a^2 & \text{for } |a| \leq \delta, \\
\delta \left( |a| - \frac{1}{2} \delta \right) & \text{otherwise},
\end{cases}
\end{equation}

\noindent where \( a = y - f(x) \). 

Let us now introduce a more specific loss constraint that is directly related to hand articulation and our problem. The \textbf{anatomical joint limits error }\( E_{\text{limits}} \) is a penalty term applied to the 45-dimensional pose vector, where each joint has three DoF. This error function ensures that estimated hand poses remain within experimentally determined anatomical constraints, thereby enforcing plausible hand articulations. Similar to \cite{inproceedings}, this loss is implemented as a soft constraint using exponential functions that activate when joint angles exceed predefined limits:
\begin{equation}
E_{\text{limits}}(\boldsymbol{\theta}) = a_{\text{limits}} \sum_{i=0}^{m} \left(e^{l_i - \theta_i} + e^{\theta_i - u_i} \right),
\end{equation}
\noindent where \( [l_i, u_i] \) denote the lower and upper bounds for joint angle \( \theta_i \), and \( a_{\text{limits}} \) is an experimentally determined weight factor. The exponential terms enforce smooth constraints that discourage hand poses outside the allowable range.
The concept of anatomical constraints for hand motion was first introduced by Lin ~\etal\cite{897381} and has since been adopted by various works, including \cite{inproceedings}, \cite{Tzionas_2016}, and others.

\textbf{Rigid Transformations.}
A rigid transformation \cite{rigidtrans} involves applying a rotation, translation, and optionally scaling to align two sets of points 
 while preserving their internal geometric relationships.
A rigid transformation consists of two main operations:

\begin{itemize}
    \item Translation: A shift of an object in 3D space along the \( x \), \( y \), and \( z \) axes without altering its orientation.
    \item Rotation: A transformation that changes the orientation of an object while preserving its shape and size.
\end{itemize}

\section{Method}
\label{sec:method}
\begin{figure}[!t]
	\includegraphics[width=\linewidth]{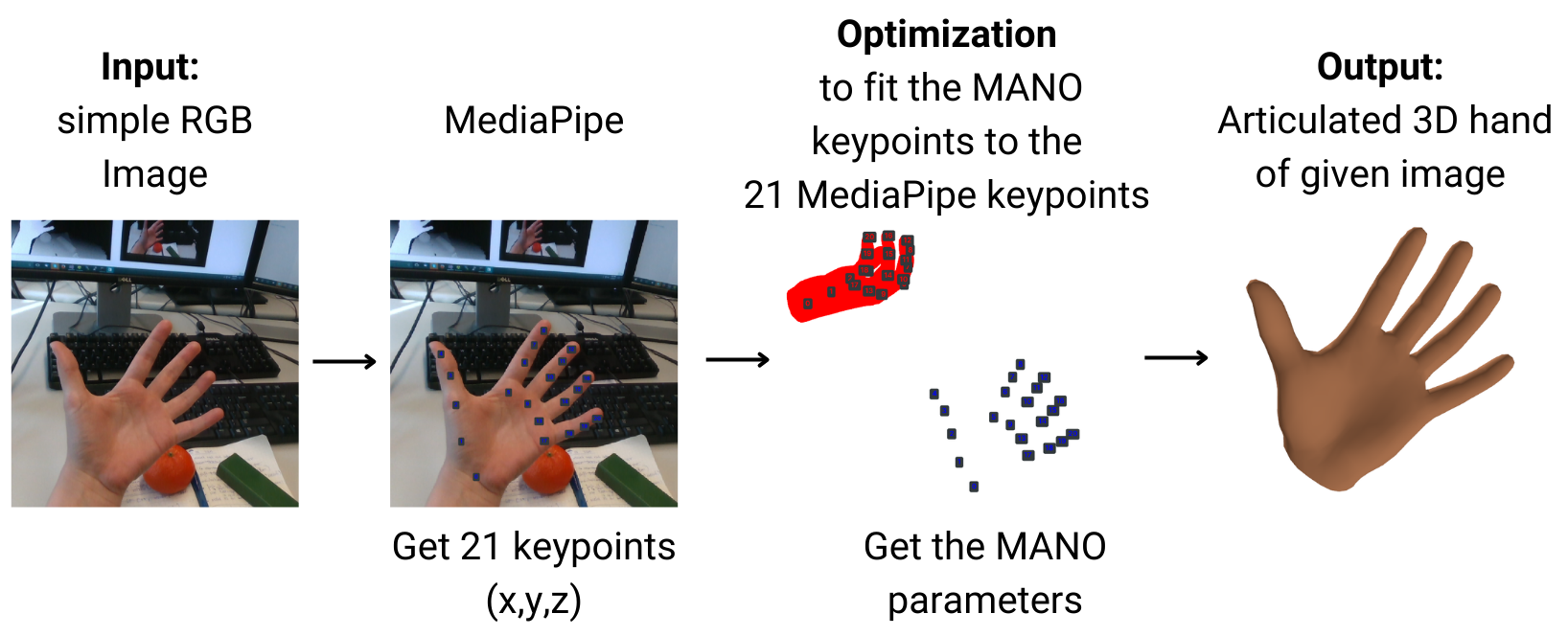}
	\centering
	\caption{Our proposed pipeline.}
	\label{pipeline}
\end{figure}
\textbf{ Overview.} To tackle the problem of fitting an articulated 3D hand from a single RGB image, we design the pipeline shown in Fig.~\ref{pipeline}.
The input is a standard RGB image containing a human hand, and the output is a 3D hand in the exact same pose and orientation.
The first step in the pipeline involves passing the input image through the Media\-Pipe Hand model estimator \cite{Mediapipehand} to extract the 21 hand keypoints along with handedness information (whether the hand is left or right). This process results in a list of keypoints corresponding to pixel locations in the input image, which we later use as ground truth parameters for fitting the MANO hand model~\cite{MANO}.
For the optimization step, our goal is to fit the MANO model’s keypoints to the extracted ``ground truth" keypoints from Media\-Pipe. Given the pose and shape parameters of MANO, we can obtain the 21 hand joint locations through linear interpolation. MANO takes 45 pose parameters, 10 shape parameters, and 3 global rotation parameters as input to generate a 3D mesh that represents a unique hand configuration.
The fitting process begins with a neutral ``zero" pose and shape. Through a series of transformations and iterative optimization, we adjust the MANO parameters to align its 21 keypoints with the estimated Media\-Pipe keypoints by minimizing a loss function. 
A key challenge was when the input hand was in a different global rotation than MANO’s default pose. 
The root of the problem was the initialization: MANO starts in a neutral pose, shape, and rotation, since we have no prior information about the input hand's orientation, and initialized every parameter with zeros. However, if the hand in the input image had a significantly different rotation, for example, in a ``handshake" position, the optimization process failed, resulting in completely implausible hand meshes.

\begin{figure}[!t]
	\includegraphics[width=\linewidth]{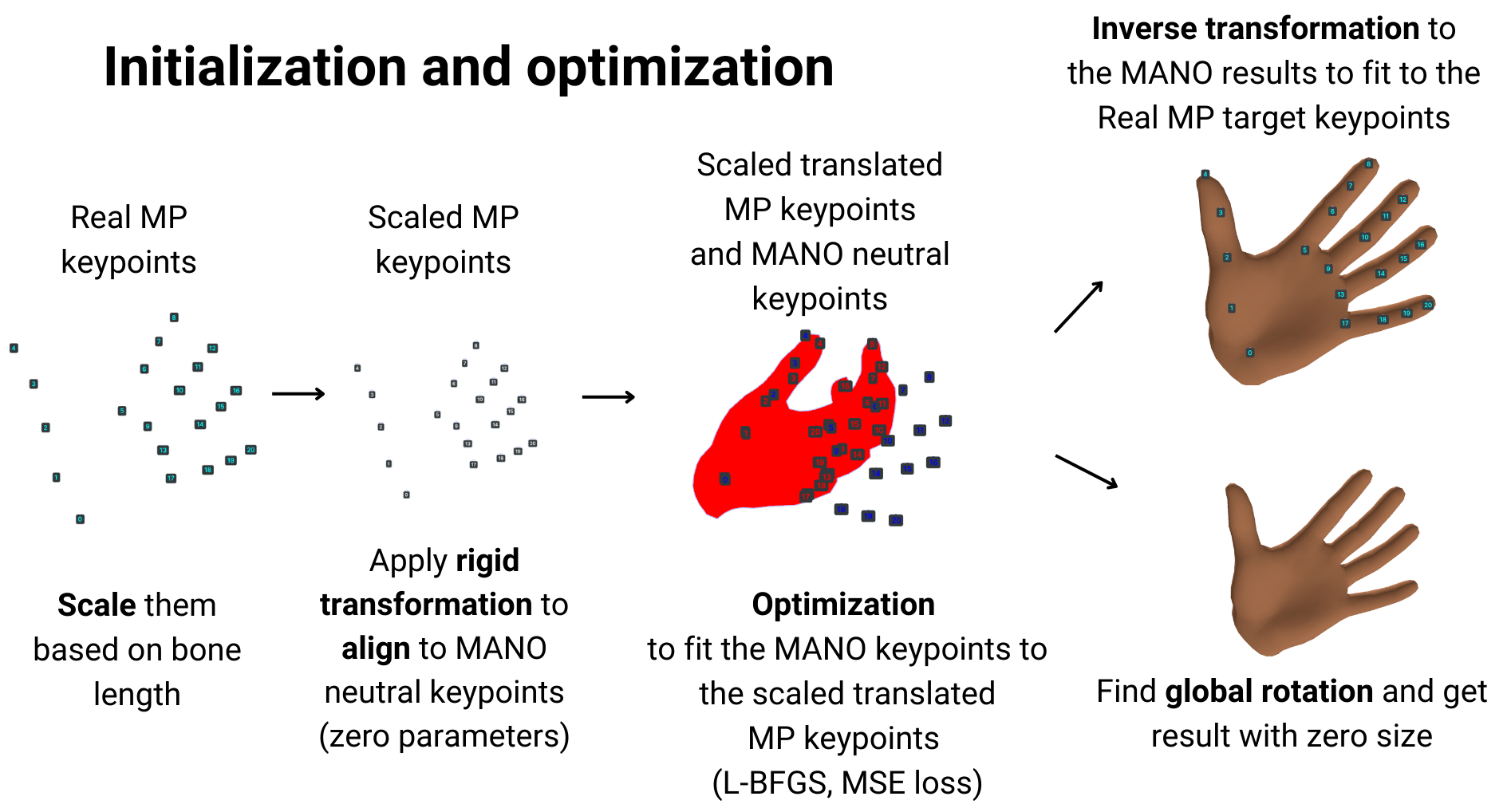}
	\centering
	\caption{Optimization pipeline (MP stands for Media\-Pipe).}
	\label{optpipeline}
\end{figure}

\begin{table*}[t]
    \centering
    \caption{The description of what combination of optimizer and loss function each experiment uses.}
    \footnotesize
    \begin{tabular}{|c|cc|ccccccc|}
    \hline
        \multirow{2}{*}{\textbf{ID}} & \multicolumn{2}{c|}{\textbf{OPTIMIZERS}} & \multicolumn{7}{c|}{\textbf{LOSSES}} \\ \cline{2-10}
        ~ & LBFGS & BFGS & MSE &\makecell{MSE\\fingertips}& \makecell{Geman-\\McClure} &\makecell{Geman-\\McClure\\fingertips} & Huber &\makecell{Huber\\fingertips} & \makecell{Anatomical\\\textbf{(2 stages)}}  \\ \hline
        \textbf{A} & \cmark & ~ & \cmark & \cmark & ~ & ~ & ~ & ~ &   \\ \hline
        \textbf{B} & \cmark & ~ & \cmark & ~ & ~ & ~ & ~ & ~ &   \\ \hline
        \textbf{C} & \cmark & ~ & ~ & ~ & \cmark & \cmark & ~ & ~ &   \\ \hline
        \textbf{D} & \cmark & ~ & ~ & ~ & \cmark & ~ & ~ & ~ &   \\ \hline
        \textbf{E} & \cmark & ~ & ~ & ~ & ~ & ~ & \cmark & \cmark &   \\ \hline
        \textbf{F} & \cmark & ~ & ~ & ~ & ~ & ~ & \cmark & ~ &   \\ \hline
        \textbf{G} & ~ & \cmark & \cmark & \cmark & ~ & ~ & ~ & ~ &   \\ \hline
        \textbf{H} & \cmark & ~ & \cmark & ~ & ~ & ~ & ~ & ~ & \cmark \\ \hline
    \end{tabular}
    \label{ablationexp}
\end{table*}

\textbf{Optimization Pipeline.}
To address this challenge, we compute a rigid transformation aligning the neutral MANO keypoints to MediaPipe keypoints.
Additionally, we applied scaling to ensure both sets of keypoints were properly aligned before beginning the optimization process. The complete optimization pipeline is illustrated in Fig.~\ref{optpipeline}.
The rigid transformation was computed using six stable palm joints, namely keypoints [0,1,5,9,13,17] as illustrated  in Fig.~\ref{bloghands} to minimize the influence of finger articulation.
To compute the transformation, we implemented a custom function that returns a $4\times4$ transformation matrix, where the top-left $3\times3$ block represents the rotation, 
the top-right $3\times1$ column is the translation vector, and the bottom row is used for homogenous coordinates.
For scaling, we computed a scale factor using the distance between keypoints 0 and 5 (wrist to index MetaCarpoPhalangeal joint) in both the target and MANO keypoints, ensuring anatomically proportional alignment.
After the initial alignment, we used scipy.minimize~\cite{scipyminimize} to optimize the MANO parameters, specifically leveraging either the BFGS, or the L-BFGS method. 
We experimented with three loss functions: MSE, GM, and Huber loss.
To improve the accuracy of fingertip alignment, we explored weighted loss functions, as we observed that most keypoints were densely concentrated around the palm. Due to the nature of the loss functions, the optimization process primarily focused on minimizing the error in these denser regions, often leading to less precise alignment of the fingertips. However, in real-world hand movement, fingertips play a critical role in defining hand gestures and poses, making their accurate positioning essential. To address this, we applied weighted loss functions across all three variations (MSE, GM, and Huber), giving higher importance to fingertip keypoints to ensure their proper alignment.
Additionally, we attempted to integrate anatomical joint constraints into the loss function to enforce physically plausible hand poses. However, this approach proved ineffective, as it restricted the optimization process too severely. Despite experimenting with various weighting schemes, the results did not improve. Instead, we adopted a two-stage optimization strategy:
\begin{itemize}
    \item Stage 1: A standard MSE loss function was used to obtain an initial estimate of the hand pose.
    \item Stage 2: The output of Stage 1 was refined using anatomical loss constraints, combined with an MSE loss applied only to 2D keypoints. This ensured that the estimated hand remained within realistic anatomical bounds.
\end{itemize}
\noindent After optimization, we reversed the rigid transformation to recover the MANO keypoints and mesh in the original scale, 
orientation, and position. The rotation matrix, translation vector, and scaling factor used for initialization were inverted 
to map the optimized MANO results back to the target’s coordinate system so that we can inspect our results visually. 
To determine the root pose, we computed the axis-angle representation of the rotation matrix derived from the rigid transformation 
and incorporated it into the MANO model parameters. This ensured that the global rotation was accurately represented in the final output. This means that we now also have a 3D hand with the correct global orientation and pose but with the zero shape that is not really a problem.

\section{Experiments}
\label{sec:experiment}
For the quantitative evaluation of our method, we follow the evaluation pipeline proposed in \cite{inproceedings} and \cite{zhou2020monocular}. This ensures direct comparability between our approach and both their methods, as well as with other SotA techniques they benchmarked.
Although \cite{inproceedings} employs an optimization-based approach similar to ours, while \cite{zhou2020monocular} follows a learning-based approach, we evaluate our method against both SotA deep learning methods to provide a comprehensive performance assessment.

\textbf{Implementation Details.}
For optimization, we employed the BFGS and L-BFGS algorithms from scipy.minimize \cite{scipyminimize}.
Regarding loss functions, we utilized PyTorch’s built-in MSELoss for Mean Squared Error computation. 
However, for the Geman-McClure and Huber loss functions, we implemented custom versions to ensure proper integration within our framework. 
Additionally, anatomical constraints are inherently implemented in the Manotorch \cite{manotorch} framework, which was a key factor in our decision to adopt it. 


\begin{table}[t]
    \centering
    \caption{Ablation Study: End-Point Error (mm) ($\downarrow$) and AUC of PCK ($\uparrow$) results of EgoDexter and Dexter+Object.}
    \begin{tabular}{|l|cc|cc|}
    \hline
        ~ & \multicolumn{2}{c|}{\textbf{EgoDexter}} & \multicolumn{2}{c|}{\textbf{Dexter+Object}} \\ \hline
        ID & \makecell{EPE\\(mm) ($\downarrow$)}& \makecell{AUC of \\PCK ($\uparrow$)} & \makecell{EPE\\(mm) ($\downarrow$)}& \makecell{AUC of \\PCK ($\uparrow$)}\\ \hline
        A & \textbf{17.724} & \textbf{0.883} & 13.985 & \textbf{0.946}  \\ \hline
        B & 19.642 & 0.859 & 14.859 & 0.943  \\ \hline
        C & 17.729 & \textbf{0.883} & 13.980 & \textbf{0.946}  \\ \hline
        D & 19.648 & 0.859 & 14.878 & 0.942  \\ \hline
        E & 17.864 & 0.882 & 13.978 & \textbf{0.946}  \\ \hline
        F & 19.963 & 0.855 & 15.279 & 0.939  \\ \hline
        G & 17.787 & \textbf{0.883} & \textbf{13.975} & \textbf{0.946}  \\ \hline
        H & 42.833 & 0.492 & 24.574 & 0.784 \\ \hline
    \end{tabular}
    \label{ablationres}
\end{table}
\textbf{Using Images in the Wild.}
With the improvements in initialization and optimization, the model is now robust enough to handle hand keypoints extracted from images captured ``in the wild," such as casual or uncontrolled environments. Moreover, our current implementation does not use the camera parameters so every 2D image containing a hand can be used as an input.

\textbf{Metrics.}
To ensure a direct and fair comparison with SotA methods, we evaluate our approach using the same metrics as \cite{inproceedings} and \cite{zhou2020monocular}. These metrics assess the accuracy of 3D hand pose estimation by measuring the deviation of predicted keypoints from the ground truth.
 
We compute the Root Mean Square Error (RMSE), also referred to as End-Point Error (EPE), for all 3D hand joints. This metric quantifies the absolute difference between the estimated keypoints and the ground truth in millimeters. A lower RMSE value indicates a more precise reconstruction of the hand pose.

The Percentage of Correct Keypoints (PCK) evaluates the proportion of keypoints that are correctly estimated within a given threshold distance from the ground truth. We compute PCK for thresholds ranging from 20mm to 50mm to analyze accuracy at different tolerance levels. To summarize the PCK performance across different thresholds, we also compute the Area Under the Curve (AUC) for this range. A higher AUC value indicates better overall accuracy, with an ideal score reaching 1.0, representing a perfect fit. This metric is particularly useful for comparing methods holistically, as it accounts for performance across multiple threshold levels rather than relying on a single fixed distance.

\textbf{Datasets.}
Since our method does not involve a training phase, we exclusively use datasets for evaluation purposes.

EgoDexter~\cite{OccludedHands_ICCV2017} is an RGB-D dataset designed for evaluating hand-tracking algorithms in cluttered environments with significant occlusions. It consists of four video sequences featuring four different actors (two female) interacting with various objects in diverse settings.
Ground truth annotations include manually labeled 3D fingertip positions on depth data. However, only fingertips are annotated, and due to occlusions, not all frames contain annotations for all five fingertips.

Dexter+Object~\cite{RealtimeHO_ECCV2016} is an RGB-D dataset designed for evaluating algorithms that track both hands and objects simultaneously. It comprises six video sequences featuring two actors (one female) interacting with a simple cuboid-shaped object. 
Ground truth annotations include manually labeled 3D fingertip positions and three cuboid corners on depth data. However, for our evaluation, we only utilize the five-fingertip positions. Although all frames contain annotations, occlusions are present, particularly in sequences in which the cuboid obstructs parts of the hand.
Both datasets provide manually annotated 3D hand keypoints in millimeters in camera coordinates. However, our method, which is based on MediaPipe’s predictions, outputs 21 keypoints in pixel space in $(x, y, z)$ format, including depth values. Thus, we must transform them to camera coordinates to ensure compatibility with the dataset’s ground truth values. 

\subsection{Ablation Study}
Tab.~\ref{ablationexp} presents the different experimental configurations we tested to evaluate our method on two datasets. Each experiment is identified by a letter (ID) and corresponds to a specific combination of optimizers and loss functions. 
While we tested all loss function combinations using the L-BFGS optimizer, we selectively tested what we considered the most promising loss function setup with the BFGS optimizer.
Tab.~\ref{ablationres} presents the quantitative results of our experiments, reporting the End-Point Error (in mm) and the AUC of PCK for both datasets. Our findings indicate that most experiments perform consistently well, with results comparable to each other.
This consistency is expected, as the different loss functions are all variations of the MSE loss, designed to improve robustness against outliers. The best-performing results for each metric are highlighted in bold, although in many cases, multiple configurations achieve similar results.
Our observations suggest that weighted fingertips loss functions tend to yield slightly better performance across metrics. However, the standard loss functions also achieve competitive results, while the better results of the weighted fingertips loss may be attributed to the specific evaluation datasets focusing primarily on fingertip keypoints.
Additionally, we can observe that in experiment H, this two-stage approach integrating anatomical constraints resulted in worse overall performance although it successfully corrected depth-related errors caused by Media\-Pipe’s 3D predictions in specific cases.

\begin{figure}[!t]
    \begin{subfigure}[b]{\linewidth}
	\includegraphics[width=0.8\linewidth]{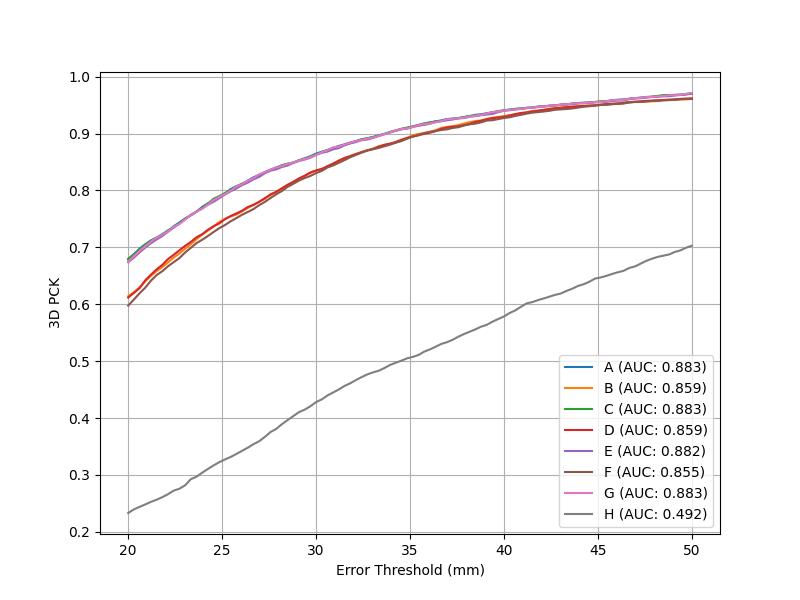}
	\centering
	\caption{3D PCK results on the EgoDexter dataset.}
	\label{pckED}
    \end{subfigure}
    
    
    \begin{subfigure}[b]{\linewidth}
	\includegraphics[width=0.8\linewidth]{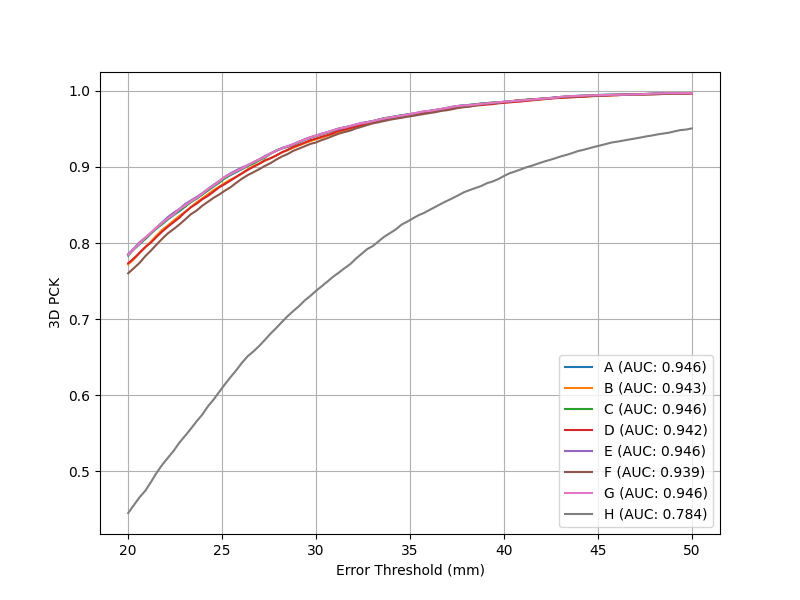}
	\centering
	\caption{3D PCK results on the Dexter+Object dataset.}
	\label{pckDO}
    \end{subfigure}
    \caption{3D PCK evaluation results.}
\end{figure}

Overall, our results suggest that the most effective optimizer-loss combinations were Experiment A and Experiment G. This outcome aligns with expectations, as L-BFGS and BFGS are closely related, with L-BFGS being a memory-efficient variant. If computational efficiency is considered alongside accuracy, Experiment A, using L-BFGS with a weighted combination of MSE losses, emerges as the optimal choice.
Figs.~\ref{pckED} and \ref{pckDO} compare the performance of the methods discussed above on the EgoDexter and Dexter+Object datasets, respectively. In both cases, the method incorporating anatomical constraints performs the worst. For the EgoDexter dataset, we observe that methods utilizing weighted fingertip loss functions consistently outperform those with standard loss functions, forming a distinct cluster of higher-performing models. However, on the Dexter+Object dataset, this distinction is less pronounced.

\begin{figure*}[!t]
  \centering
  \begin{subfigure}{0.48\linewidth}
    \includegraphics[width=0.8\linewidth]{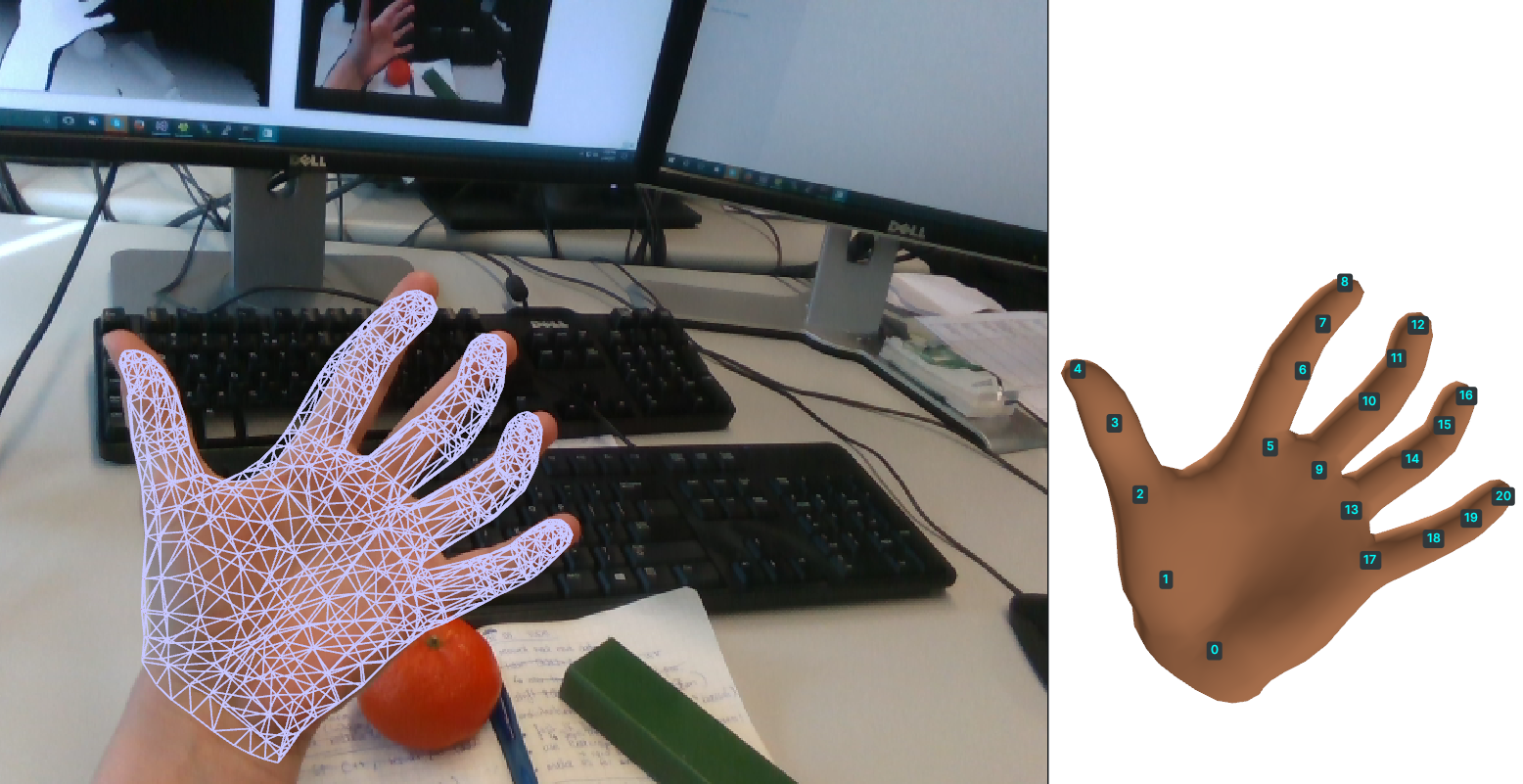}
    \caption{Simple hand pose from EgoDexter.}
    \label{ressimple}
  \end{subfigure}
  \hfill
  \begin{subfigure}{0.48\linewidth}
    \includegraphics[width=0.8\linewidth]{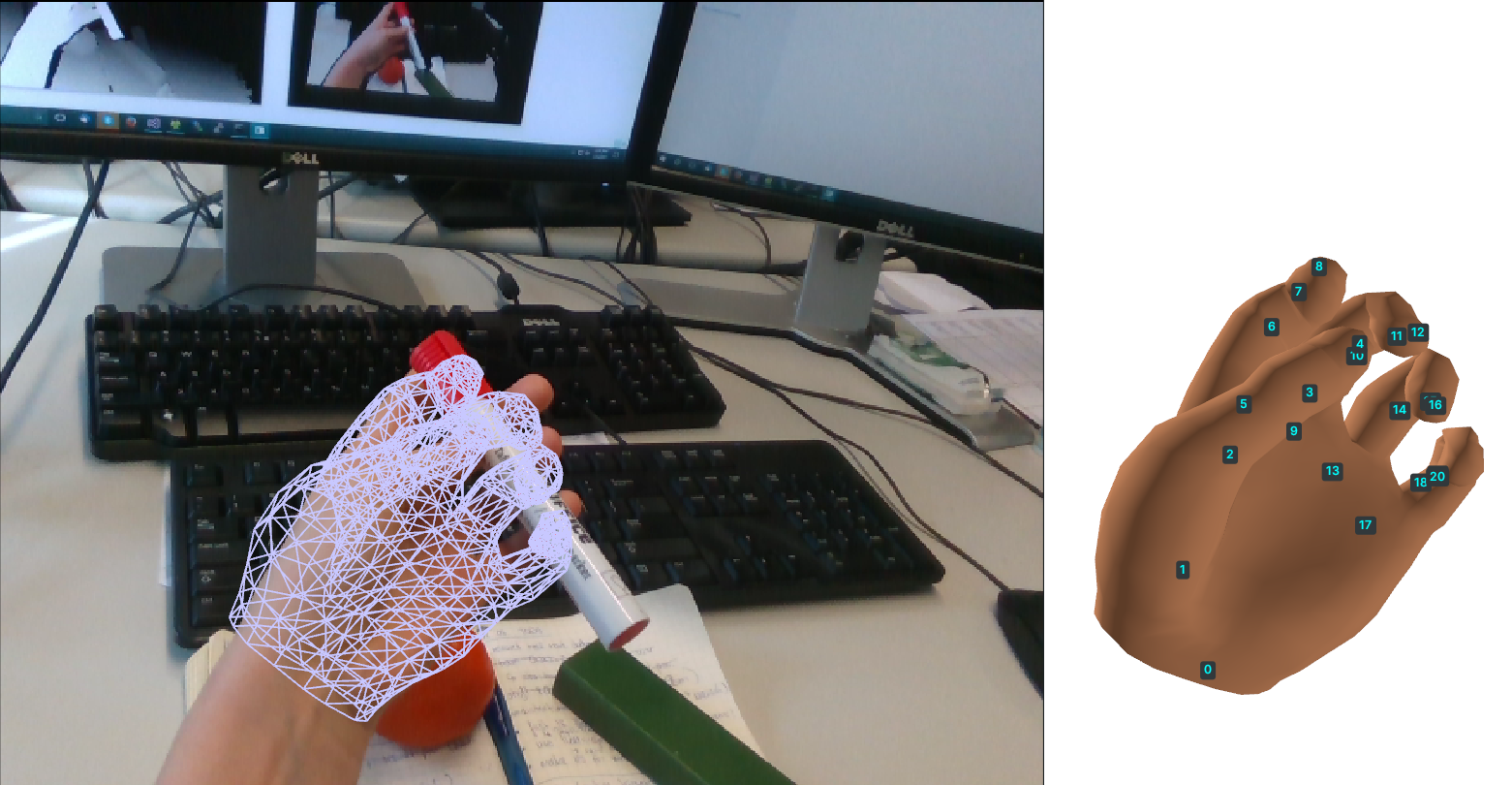}
    \caption{Challenging hand pose with occlusion from EgoDexter.}
    \label{reschal}
  \end{subfigure}
  \begin{subfigure}{0.48\linewidth}
    \includegraphics[width=0.8\linewidth]{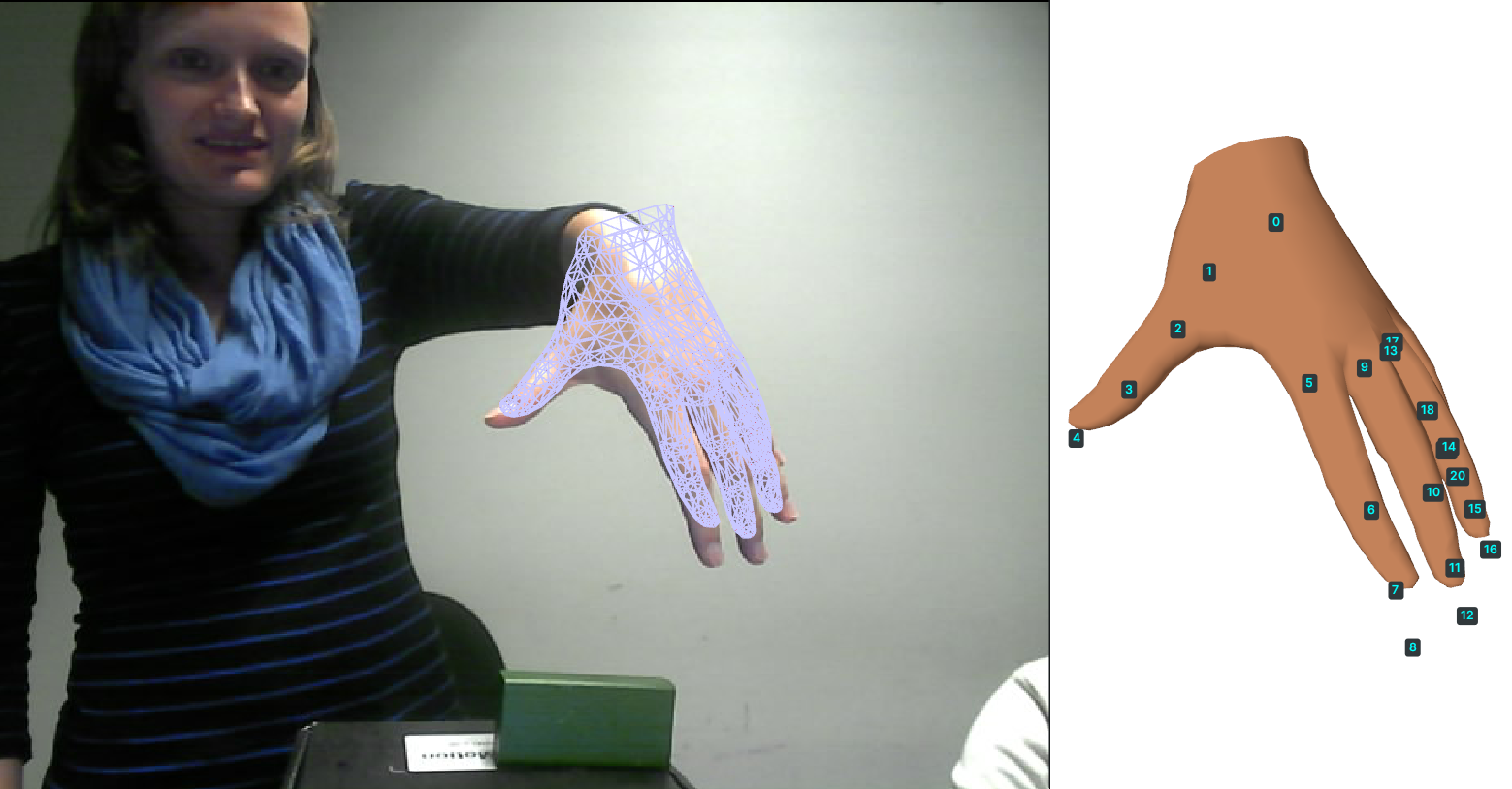}
    \caption{Dexter+Object result with MSE loss. Fingertip misalignment.}
    \label{do1bad}
  \end{subfigure}
  \hfill
  \begin{subfigure}{0.48\linewidth}
    \includegraphics[width=0.8\linewidth]{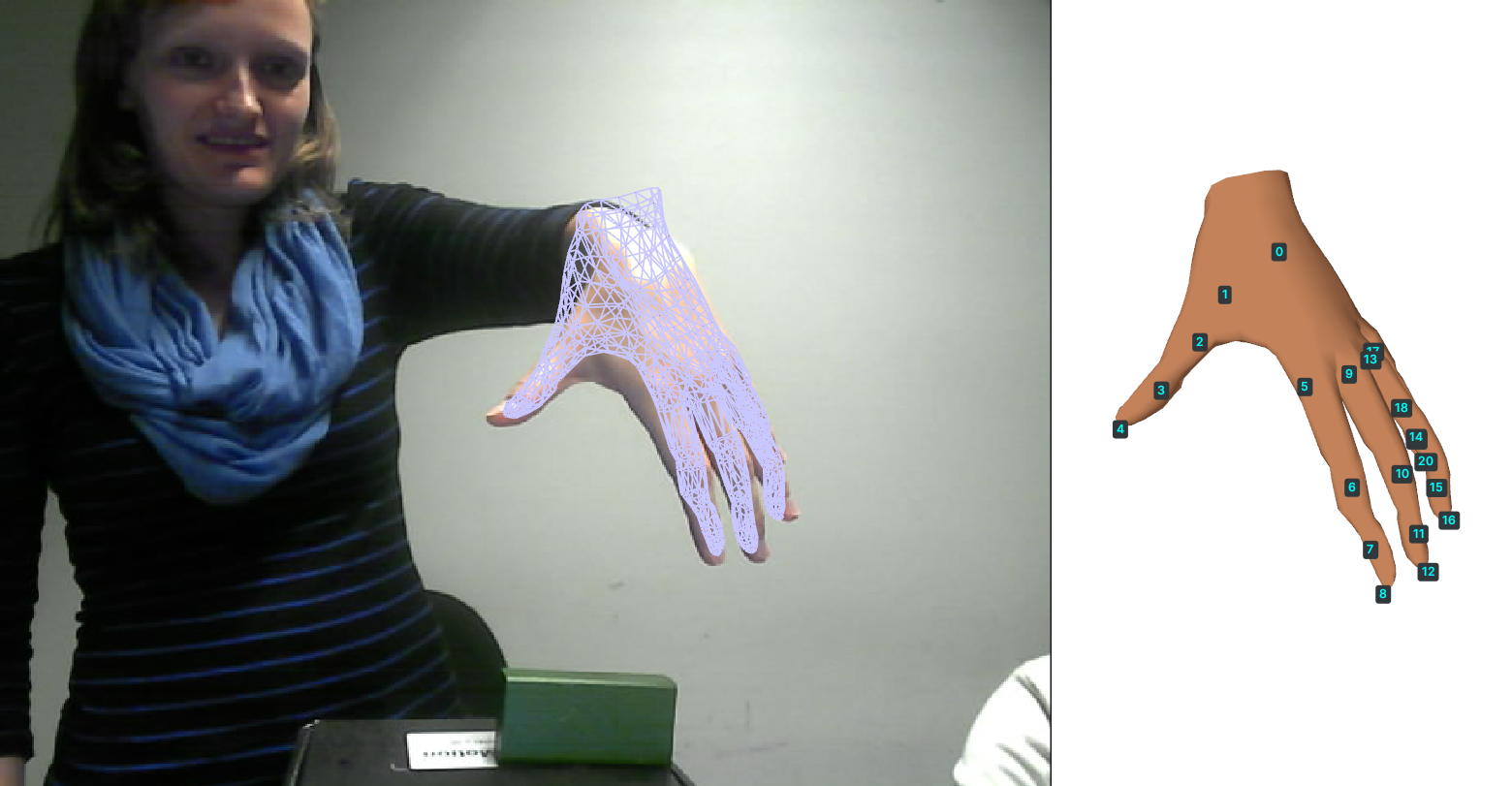}
    \caption{Same image with weighted fingertips MSE loss. Fingertip alignment improves.}
    \label{do1good}
  \end{subfigure}
  \begin{subfigure}{0.48\linewidth}
    \includegraphics[width=0.5\linewidth]{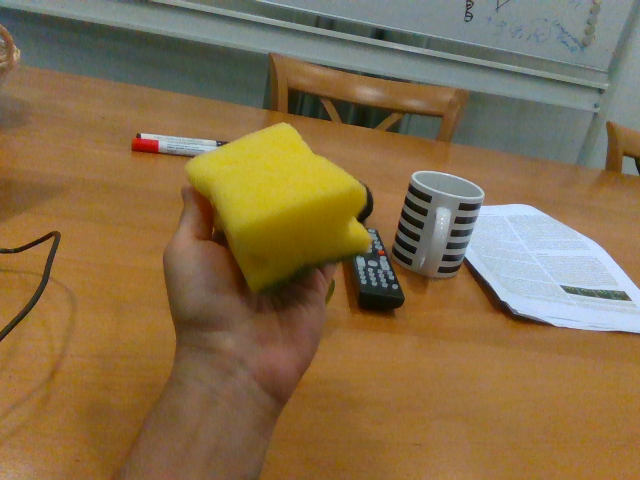}
    \caption{Failure case due to severe occlusion.}
    \label{fail}
  \end{subfigure}
  \hfill
  \begin{subfigure}{0.48\linewidth}
    \includegraphics[width=0.6\linewidth]{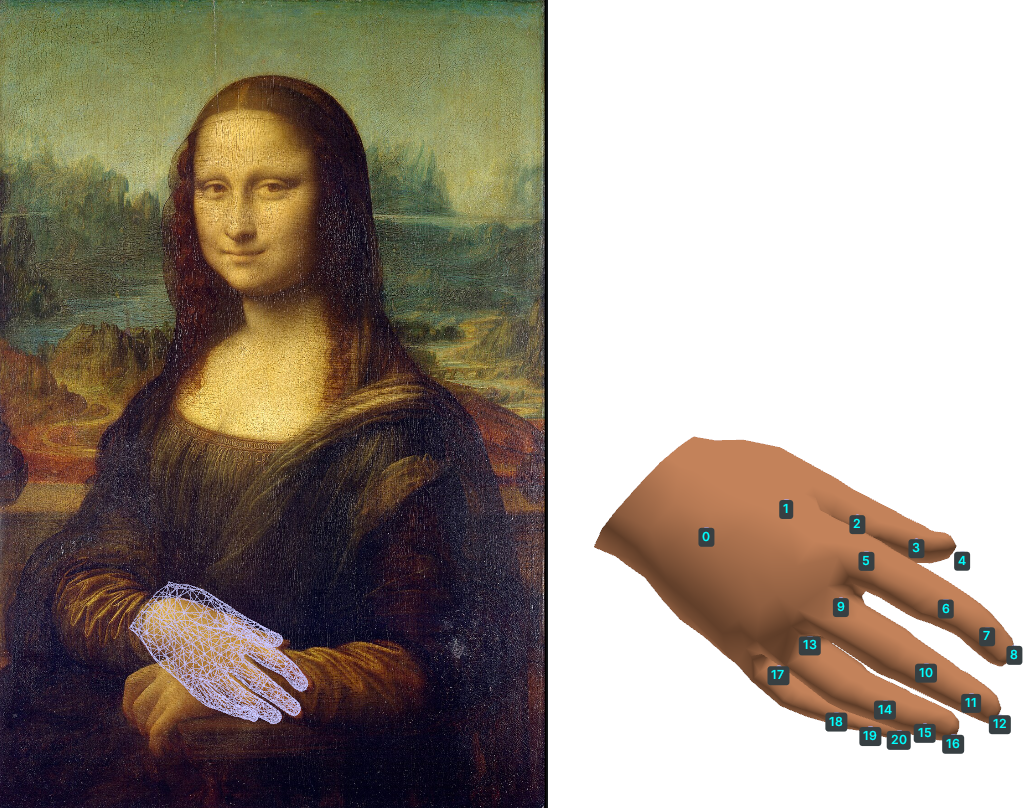}
    \caption{In-the-wild test on the Mona Lisa painting. Method generalizes well.}
    \label{monalisa}
  \end{subfigure}

  \caption{Qualitative results on both EgoDexter and Dexter+Object datasets as well as images ``in the wild." Examples include successful and failure cases, demonstrating both robustness and limitations of the method.}
  \label{fig:qualresults}
\end{figure*}
\subsection{Quantitative Evaluation}
\begin{table}[!t]
    \centering
    \caption{AUC of PCK ($ \uparrow $) comparison with state-of-the-art methods. We use ``*” to note the methods that work in real-time, which is a more challenging task.}
    \begin{tabular}{|l|c|c|}
    \hline
        Method & \textbf{\makecell{Dexter+\\Object}} & \textbf{EgoDexter} \\ \hline
        \textbf{Ours} & 0.946 & \textbf{0.883}  \\ \hline
        Zhou~\etal \cite{zhou2020monocular} & \textbf{0.948*} & 0.811*  \\ \hline
        Zhang~\etal \cite{zhang2019endtoend} & 0.825 &  - \\ \hline
        Baek~\etal \cite{baek2019pushing} & 0.650 &  - \\ \hline
        Xiang~\etal \cite{Xiang2018} & 0.912 &  - \\ \hline
        Boukhayma~\etal \cite{boukhayma20193d} & 0.763 & 0.674  \\ \hline
        Iqbal~\etal \cite{iqbal2018hand} & 0.672 & 0.543  \\ \hline
        Spurr~\etal \cite{Spurr} & 0.511 & -  \\ \hline
        Mueller~\etal \cite{GANeratedHands_CVPR2018} & 0.482* &  - \\ \hline
        Zimmermann and Brox \cite{zimmermann2017learning} & 0.573 &  - \\ \hline
        Drosakis and Argyros \cite{inproceedings} & 0.764 & 0.563 \\ \hline
    \end{tabular}
    \label{compresults}
\end{table}

Tab.~\ref{compresults} presents a comparison of our proposed method against SotA approaches, reporting the AUC of PCK on both the Dexter+Object and EgoDexter datasets. The table includes both optimization-based and learning-based methods.
Our method achieves the highest AUC on the EgoDexter dataset, outperforming all other approaches. For the Dexter+Object dataset, our method is nearly on par with the best-performing approach, with only a marginal difference of 0.002 compared to Zhou ~\etal\cite{zhou2020monocular}.
Notably, our method not only surpasses other optimization-based methods such as Drosakis and Argyros \cite{inproceedings} and Boukhayma~\etal \cite{boukhayma20193d}, but it also outperforms several learning-based methods. This is particularly significant since learning-based methods typically require substantial computational resources for training.
Furthermore, by comparing Tab.~\ref{compresults} with Tab.~\ref{ablationres}, we observe that not only does our best-performing configuration achieve SotA results, but even several of our alternative setups remain competitive with the top methods in the field.

Finally, it is important to emphasize that the datasets used for evaluation were not included in the training phase of any learning-based methods, ensuring a fair comparison with SotA approaches.

\subsection{Qualitative Evaluation}
To further assess the performance of our method, we present qualitative results across various scenarios in Figs.~\ref{ressimple}–\ref{monalisa}.
Figs.~\ref{ressimple} and \ref{reschal} showcase examples from the EgoDexter dataset. In the simpler case (Fig.~\ref{ressimple}), our method accurately predicts the hand keypoints. However, in the more challenging scenario (Fig.~\ref{reschal}), where occlusion from object interaction occurs, the method still performs reasonably well, though minor inaccuracies appear.
Figs.~\ref{do1bad} and \ref{do1good} compare the effect of different loss functions on the Dexter+Object dataset. The simple MSE loss (Fig.~\ref{do1bad}) results in less accurate predictions, particularly in fingertip locations. By contrast, using a weighted MSE loss with fingertip emphasis (Fig.~\ref{do1good}) improves the prediction.
In Fig.~\ref{fail}, we present a failure case caused by extreme occlusion. The EgoDexter dataset does not provide ground-truth keypoints for such cases, making evaluation difficult. Additionally, MediaPipe fails to detect a hand in this scenario, which directly impacts our method, as we rely on its initial keypoint predictions rather than ground-truth annotations from a dataset. However, in less challenging cases where MediaPipe successfully detects a hand, our approach remains effective.

Finally, in Fig.~\ref{monalisa}, we demonstrate that our method generalizes beyond structured datasets by estimating hand poses from an image of the Mona Lisa painting. This showcases that our approach does not rely on camera parameters and can function on in-the-wild RGB images, making it applicable in diverse real-world scenarios.

\section{Conclusion}
\label{sec:conclusion}
In this paper, we proposed an optimization-based solution for estimating the 3D articulation of a human hand from a single RGB image, without knowledge of the camera intrinsic parameters.
Our method leveraged the MediaPipe keypoint detector to obtain an initial estimation of the hand joints in the 2D space, and it performed a fitting stage using the MANO parametric model to obtain the 3D joint rotations. For the fitting stage we incorporated a fingertip alignment loss coupled with anatomical constraints. Our extensive evaluation demonstrated that our approach can robustly operate in-the-wild without the need for prior camera parameter information, while being  competitive when compared to SotA data-driven models.

{\small
\bibliographystyle{ieee_fullname}
\bibliography{egbib}
}

\end{document}